\documentclass{article}
\pdfoutput=1

\PassOptionsToPackage{numbers}{natbib}

\usepackage[final]{nips_2016}

\usepackage[utf8]{inputenc} 
\usepackage[T1]{fontenc}    
\usepackage{subcaption}
\usepackage{subfig}
\usepackage{url}            
\usepackage{booktabs}       
\usepackage{graphicx} 
\usepackage{amsfonts}       
\usepackage{amsmath}       
\usepackage{nicefrac}       
\usepackage{microtype}      
\usepackage{bbm}
\usepackage{todonotes}
\usepackage{multirow}
\usepackage{booktabs}

\newcommand{\presec}{\vspace{-.11in}}
\newcommand{\postsec}{\vspace{-.11in}}

\newcommand{\precap}{\vspace{-.05in}}
\newcommand{\postcap}{\vspace{-.19in}}

\def\Dist{\mathcal{D}}

\title{Nothing Else Matters: Model-Agnostic Explanations By Identifying Prediction Invariance}

\author{
  Marco Tulio Ribeiro\\
  University of Washington\\
  Seattle, WA 98105\\
  \texttt{marcotcr@cs.uw.edu} \\
  \And
  Sameer Singh\\
  University of California, Irvine\\
  Irvine, CA 92697\\
  \texttt{sameer@uci.edu} \\
  \And
  Carlos Guestrin\\
  University of Washington\\
  Seattle, WA 98105\\
  \texttt{guestrin@cs.uw.edu} \\
}

\begin{document}

\maketitle

\section{Introduction}
At the core of interpretable machine learning is the question of whether humans are able to make accurate predictions about a model's behavior. 
Assumed in this question are three properties of the interpretable output: \emph{coverage}, \emph{precision}, and \emph{effort}.
\emph{Coverage} refers to \emph{how often} humans think they can predict the model's behavior, \emph{precision} to how accurate humans are in those predictions, and \emph{effort} is either the up-front effort required in interpreting the model, or the effort required to make predictions about a model's behavior. 

One approach to interpretable machine learning is designing inherently interpretable models.
Visualizations of these models usually have perfect coverage, but there is a trade-off between the accuracy of the model and the effort required to comprehend it - especially in complex domains like text and images, where the input space is very large, and accuracy is usually sacrificed for models that are compact enough to be comprehensible by humans.
Experiments usually involve showing humans these visualizations, and measuring human precision when predicting the model's behavior on random instances, and the time (effort) required to make those predictions \cite{Huysmans, bayesian_case_model, decision_sets}.

\begin{figure}
    \centering
  \begin{subfigure}[b]{0.3\textwidth}
      \scriptsize
      \begin{tabular}{cc}
        \hline
        Feature & Value \\ \hline
        Age & $37 < $ Age $\leq 48$ \\
        Workclass & Private \\
        Education & $\leq$ High School \\
        Marital Status & Married \\
        Occupation & Craft-repair\\
        Relationship & Husband\\
        Race & Black\\
        Sex & Male\\
        Capital Gain & $0$\\
        Capital Loss & $0$\\
        Hours per week & $\leq 40$\\
        Country & United States \\
        \hline
      \end{tabular}
  \caption{Instance}
  \label{adult_instance}
  \end{subfigure}
  \begin{subfigure}[b]{0.31\textwidth}
      \centering
      \includegraphics[width=0.6\textwidth]{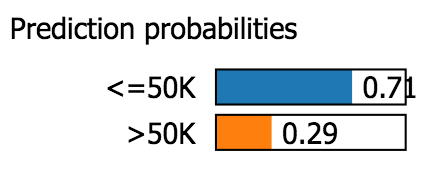}
    \includegraphics[width=\textwidth]{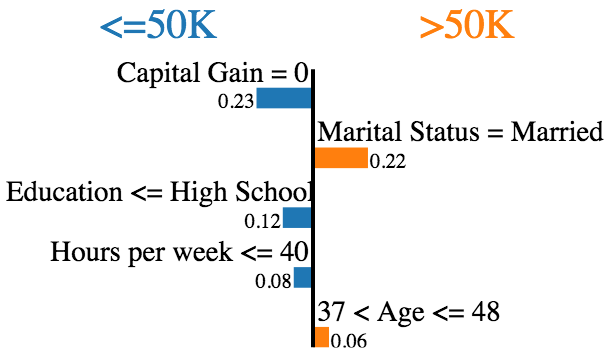}
    \caption{Linear LIME explanation}
    \label{adult_lime}
    \end{subfigure}
    \quad
  \begin{subfigure}[b]{0.35\textwidth}
      \centering
      \small
      \raisebox{15mm}{
    \begin{tabular}{|l|}
      \hline
      \textbf{IF} Education $\leq$ High School\\
      \textbf{THEN PREDICT} Salary $\leq 50K$\\
      \hline
    \end{tabular}
    }
    \caption{aLIME explanation (\emph{anchor})}
    \label{adult_alime}
    \end{subfigure}
  \caption{Explaining a prediction from the UCI adult dataset. The task is to predict if a person's salary is higher than 50,000 dollars ($>$50k) or not ($\leq$50K).}
  \label{adult_fig}
\end{figure}

Model-agnostic explanations \cite{model_agnostic} avoid the need to trade off accuracy by treating the model as a black box. Explanations such as sparse linear models \cite{lime} (henceforth called linear LIME) or gradients \cite{Baehrens:2010:EIC:1756006.1859912, deeptaylor} can still exhibit high precision and low effort (which are de-facto requirements, as there is little point in explaining a model if explanations lead to poor understanding or are too complex) even for very complex models by providing explanations that are local in their scope (i.e. not perfect coverage).
However, the coverage of such explanations are not explicit, which may lead to human error.
Take the example on Figure~\ref{adult_fig}: we explain a prediction of a complex model, which predicts that the person described by Figure~\ref{adult_instance} makes less than \$50K.
The linear LIME explanation (Figure~\ref{adult_lime}) sheds some light into why, but it is not clear whether we can apply the insights from this explanation to other instances.
In other words, even if the explanation is faithful locally, it is not easy to know what that local region is.
Furthermore, it is not clear when the linear approximation is more or less faithful, even within the local region.

In this paper, we introduce Anchor Local Interpretable Model-Agnostic Explanations (aLIME), a system that explains individual predictions with if-then rules (similar to \citet{decision_sets}) in a model-agnostic manner.
Such rules are intuitive to humans, and usually require low effort to comprehend and apply.
In particular, an aLIME explanation (or an \emph{anchor}) is a rule that sufficiently ``anchors'' a prediction -- such that changes to the rest of the instance do not matter (with high probability).
For example, the anchor in Figure \ref{adult_alime} states that the model will almost always predict Salary $\leq 50K$ if a person is not educated beyond high school, regardless of the other features.
Such explanations make their coverage very clear - they only apply when the conditions in the rule are met. 
We propose a method to compute such explanations that guarantees high precision with a high probability.
Further, we present empirical comparison against linear LIME and qualitative evaluation on a variety of tasks (such as text/image classification and visual question answering) to demonstrate that anchors are intuitive, have high precision, and very clear coverage boundaries.

\section{Anchors as Model-Agnostic Explanations}
Let the model being explained be denoted $f$, such that we explain individual predictions $f(x) = y$.
Let an anchor $c \in C_x$ be defined as a set of constraints 
(i.e. a rule with conjunctions), $C_x$ being the set of all possible constraints that are met by $x$.
For example, in Figure \ref{adult_alime}, $c = \{ \text{Education $\leq$ High School}\}$.

We assume we have a distribution of interest $\Dist$, and that we can sample from $\Dist(z | c, x)$ - that is, we can sample inputs where the constraints for the anchor $c$ are met.
The reason to condition on $x$ is that $\Dist$ may depend on the instance being explained (for example, see \emph{image classification} in Section \ref{sec:case}). The precision of an anchor is then defined as the expected accuracy (under $\Dist$) of applying the anchor to instances that meet its constraints, formalized in Equation \ref{prec_equation}.
%
\begin{equation}
  \text{Precision}(f, x, c, \Dist) = E_{\Dist(z | c, x)}[\mathbbm{1}_{f(x) = f(z)}]
  \label{prec_equation}
\end{equation}
%
As argued before, high precision is a requirement of model-agnostic explanations. It is trivial to get a perfectly precise (yet useless) anchor by having the constraint set be so specific that only the example being explained meets it. In order to balance precision, coverage and effort, we optimize the objective in Equation \ref{anchor_problem}, where we try to find the shortest anchor with high precision.
The length of the anchor can be used as a proxy for effort, and more specific (longer) anchors will naturally have less coverage.
\begin{equation}
  \begin{aligned}
    & \min_{c \subseteq C_x} & &  |c|
    & s.t. & & \text{Precision}(f, x, c, \Dist) \geq 1 - \epsilon
  \end{aligned}
  \label{anchor_problem}
\end{equation}
%
\textbf{Algorithm:} Solving Equation \ref{anchor_problem} exactly is unfeasible -- precision cannot be computed exactly for arbitrary $\Dist$ and $f$, and finding the best $c$ has combinatorial complexity.
To address the former, we approximate the precision via sampling, and solve the probably approximately correct (PAC) version of Equation~\ref{anchor_problem} so that the chosen anchor will have high precision with high probability.
For the latter, we employ an algorithm similar in spirit
to lazy decision trees \cite{lazy_dt}, where we construct $c$ greedily.
%
In particular, at each step, we want to pick the constraint that dominates all other constraints in terms of precision, until the stopping criterion in Equation \ref{anchor_problem} is met.
For efficiency, we want to sample as few instances as possible to make each greedy decision.
We use Hoeffding bounds \cite{hoeffding} for the differences in precision to decide when a constraint dominates all the other constraints with high probability.
This uses the same insight as Hoeffding trees \cite{hoeffding_dt}, with the key difference that we can control the sampling distribution, and thus can use the bounds to sample the regions of the input space that reduce the uncertainty between the precision estimates with as few samples as possible.
Due to lack of space, we omit the details of the algorithm. 

\presec
\section{Simulated Experiments}
\postsec
In order to evaluate the difference between linear LIME and anchor LIME (aLIME) in terms of coverage and precision, we perform simulated experiments on two UCI datasets: \emph{adult} and \emph{hospital readmission}. 
The latter is a 3-class classification problem, where the task is to predict if a patient will be readmitted to the hospital after an inpatient encounter within 30 days, after 30 days, or never.

For each dataset, we learn a gradient boosted tree classifier with $400$ trees, and generate explanations for instances in the validation dataset. We then evaluate the coverage and precision of these explanations on a separate test dataset. We use $\epsilon = 0.05$ for anchor (that is, we expect precision to be close to $95$\%) unless noted otherwise, and consider that a linear LIME explanation covers every other instance within distance $\tau$, where $\tau$ is a parameter that we vary.
For aLIME, we sample from $\Dist(Z | c, x)$ by sampling whole rows from the dataset except for the features constrained by $c$.
We evaluate $K$ explanations, chosen either at random (RP) or via Submodular Pick (SP), a procedure that picks explanations to maximize the coverage~\cite{lime}, on the validation. 

\begin{figure}[tb]
    \makebox[\linewidth][l]{
  \begin{subfigure}[b]{0.3\textwidth}
    \includegraphics[width=\textwidth]{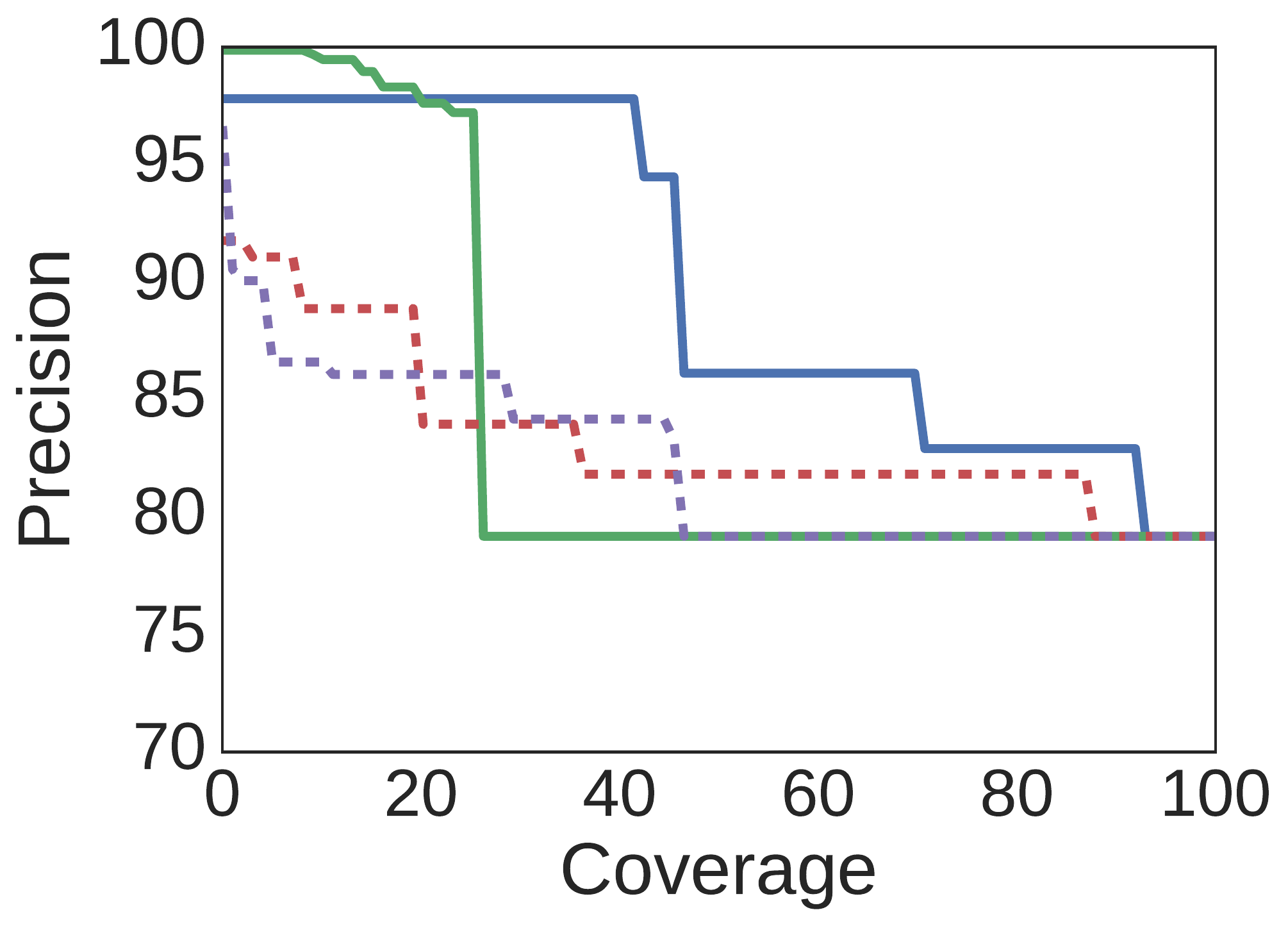}
    \caption{Precision-coverage, $K=1$}
    \label{adultpreccov}
  \end{subfigure}
  \begin{subfigure}[b]{0.3\textwidth}
    \includegraphics[width=\textwidth]{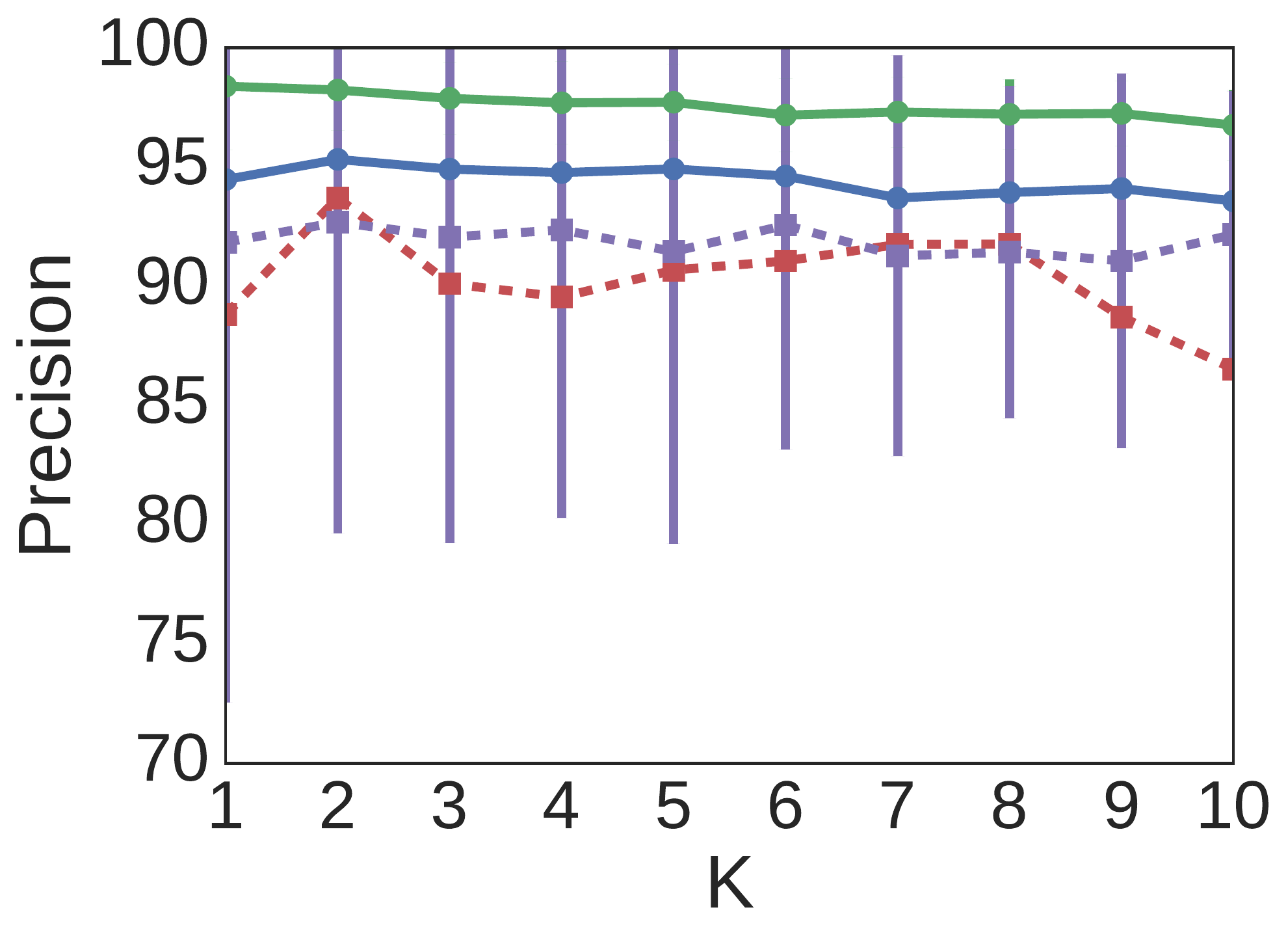}
    \caption{Precision with varying $K$}
    \label{adultprec}
  \end{subfigure}
  \begin{subfigure}[b]{0.3\textwidth}
    \includegraphics[width=\textwidth]{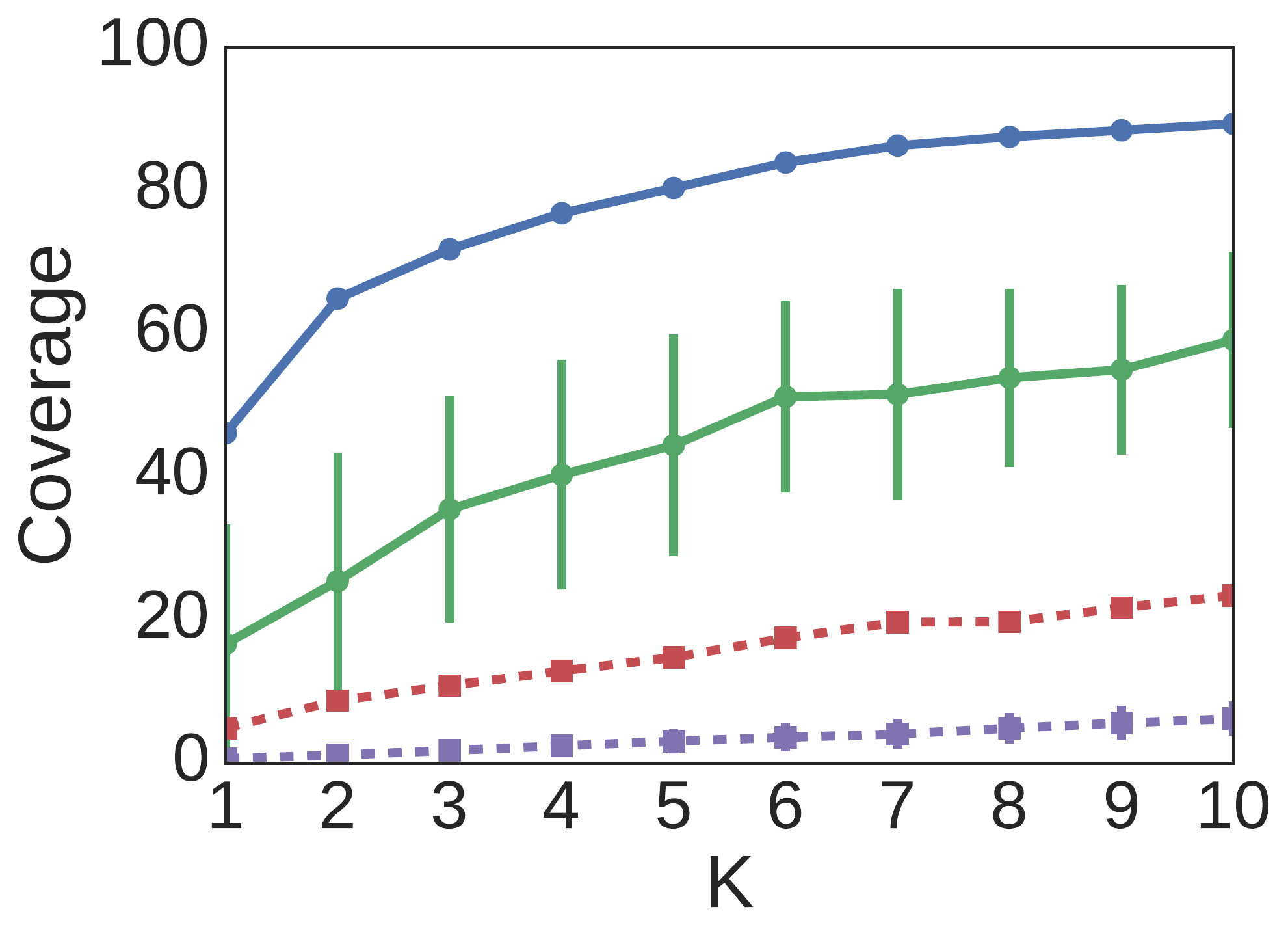}
    \caption{Coverage with varying $K$}
    \label{adultcov}
  \end{subfigure}
  \begin{subfigure}[t]{0.3\textwidth}
    \includegraphics[width=\textwidth]{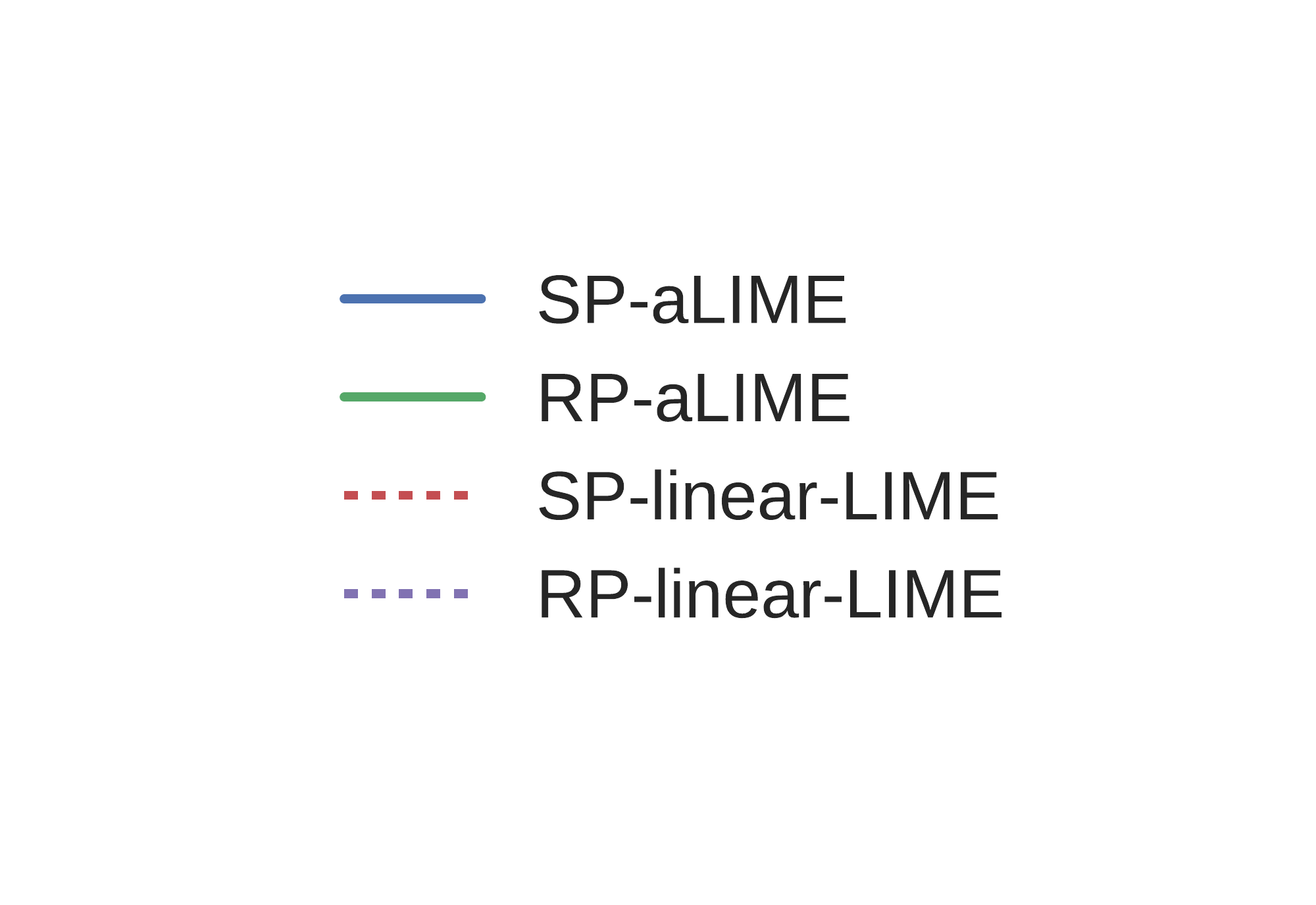}
  \end{subfigure}
  }
\caption{Adult dataset}
\label{adult-graphs}
\end{figure}

\begin{figure}[tb]
    \makebox[\linewidth][l]{
  \begin{subfigure}[b]{0.3\textwidth}
    \includegraphics[width=\textwidth]{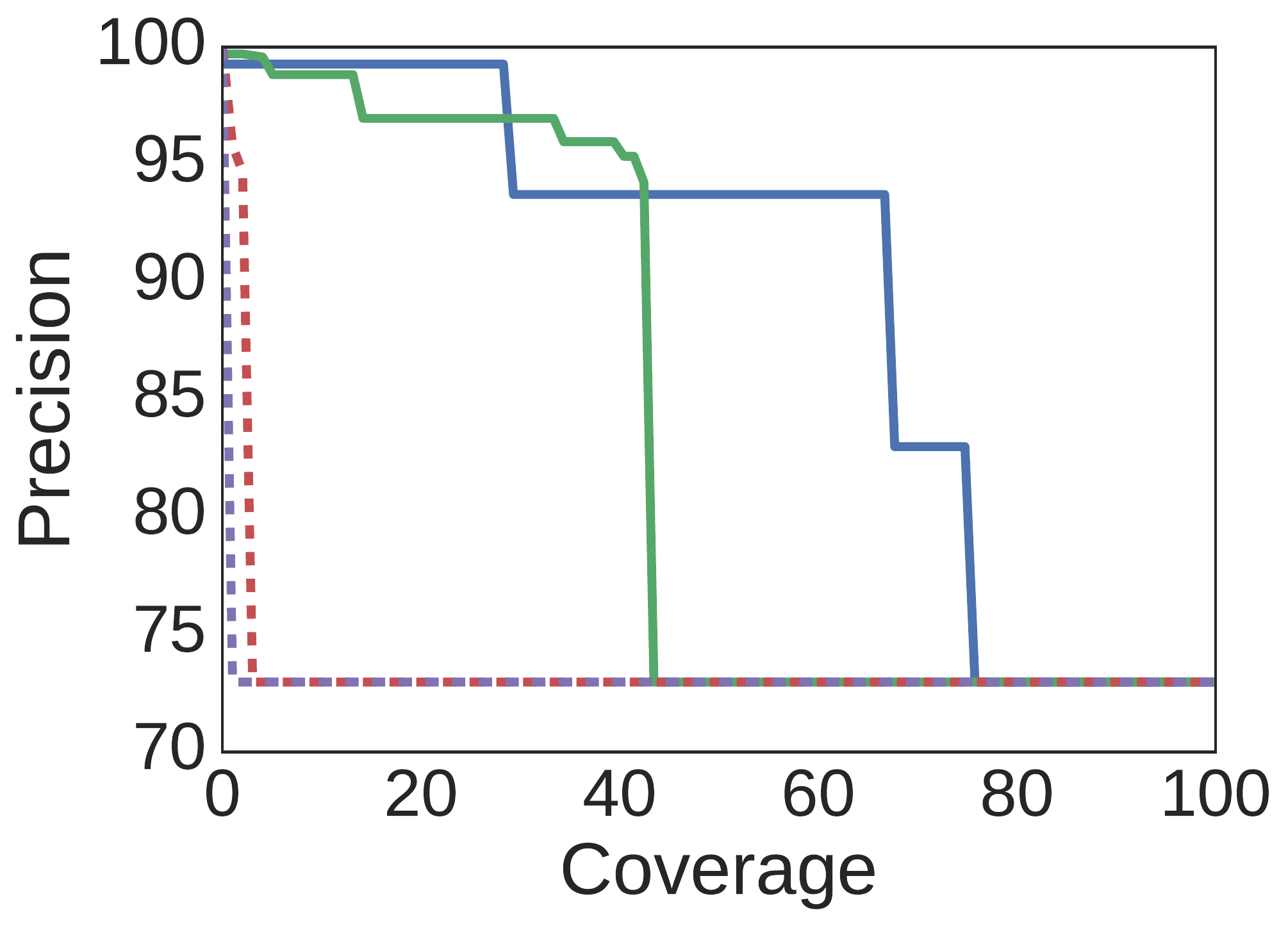}
    \caption{Precision-coverage, $K=1$}
    \label{diabetespreccov}
  \end{subfigure}
  \begin{subfigure}[b]{0.3\textwidth}
    \includegraphics[width=\textwidth]{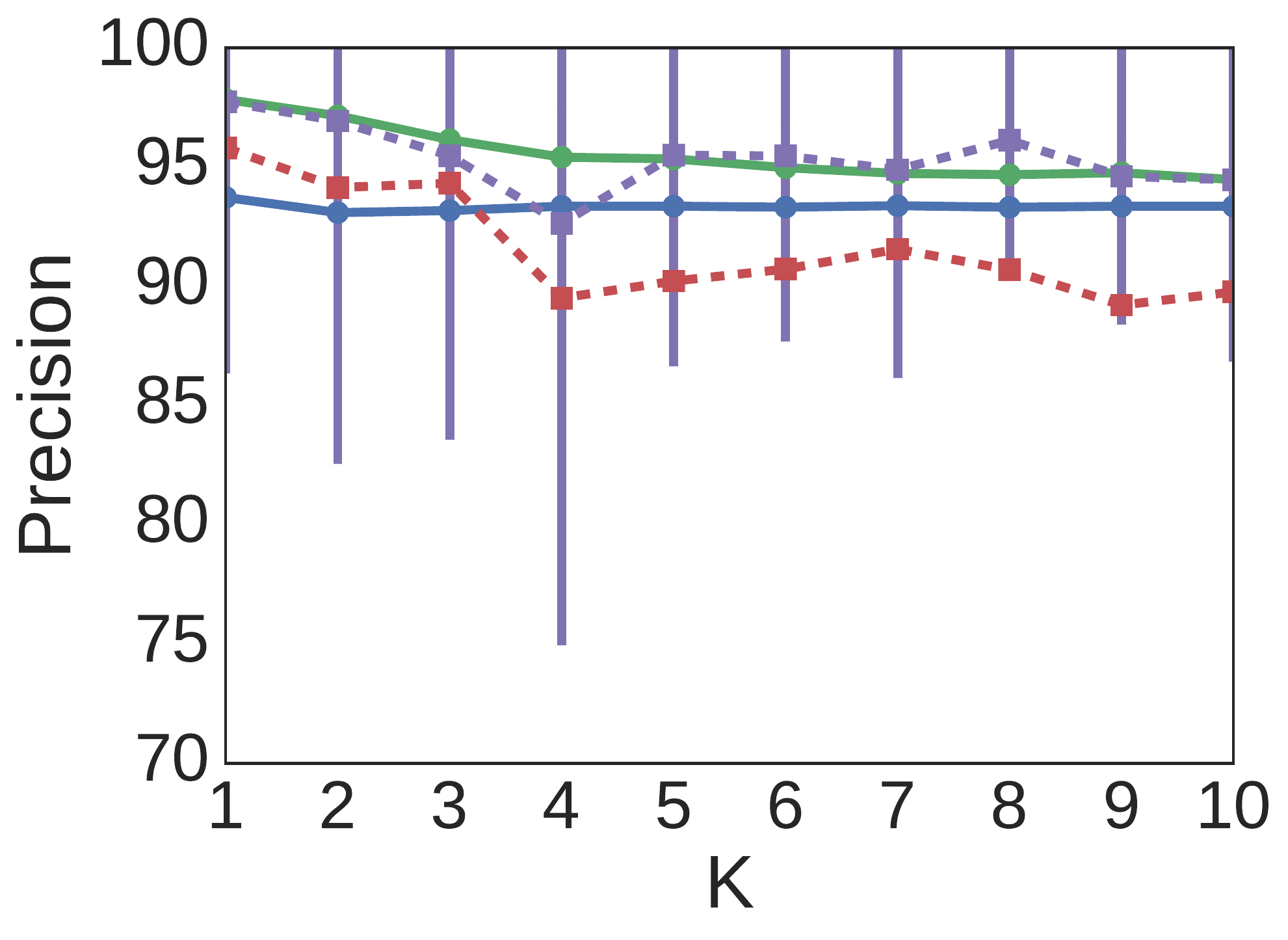}
    \caption{Precision with varying $K$}
    \label{diabetesprec}
  \end{subfigure}
  \begin{subfigure}[b]{0.3\textwidth}
    \includegraphics[width=\textwidth]{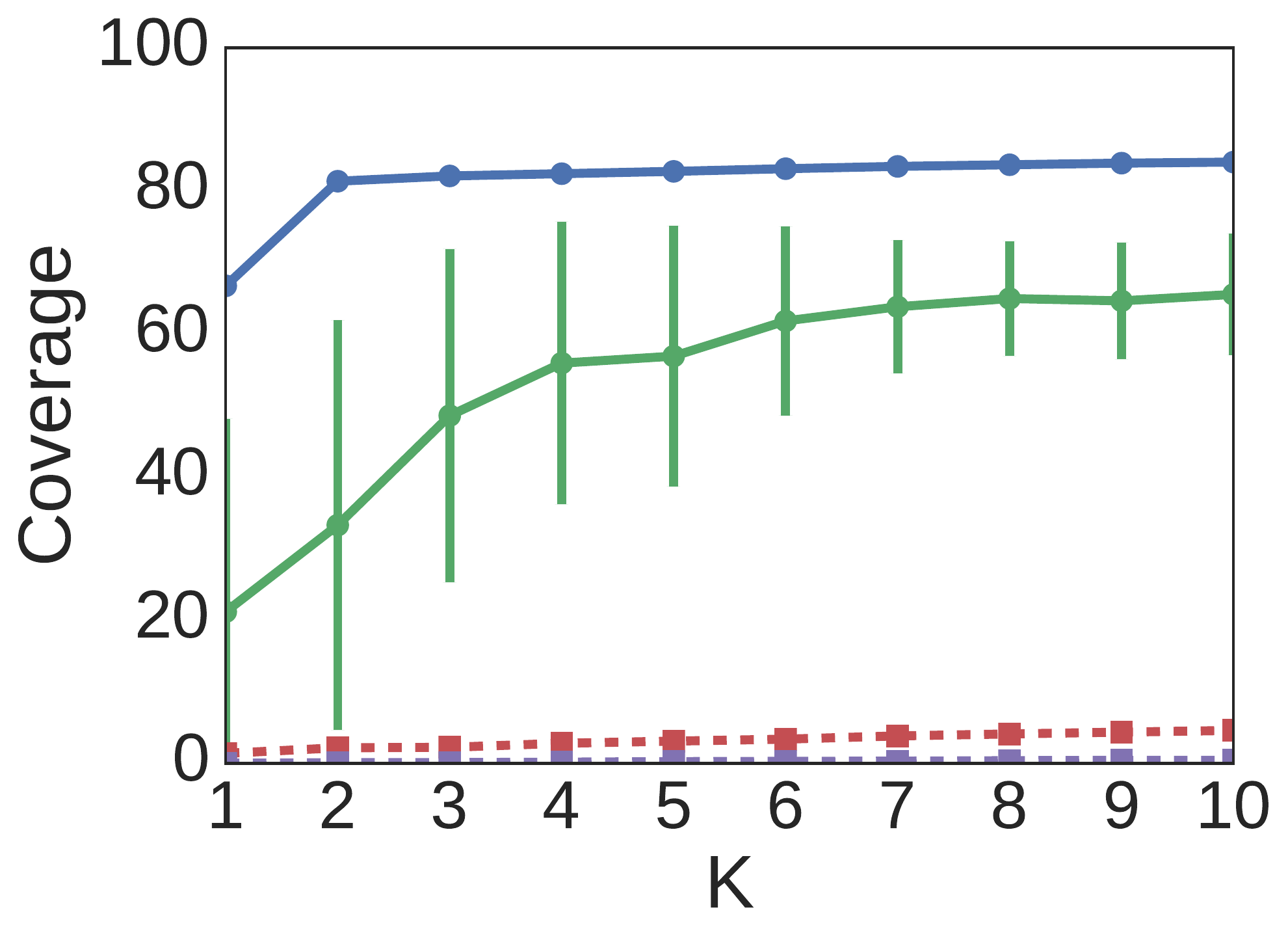}
    \caption{Coverage with varying $K$}
    \label{diabetescov}
  \end{subfigure}
  \begin{subfigure}[t]{0.3\textwidth}
    \includegraphics[width=\textwidth]{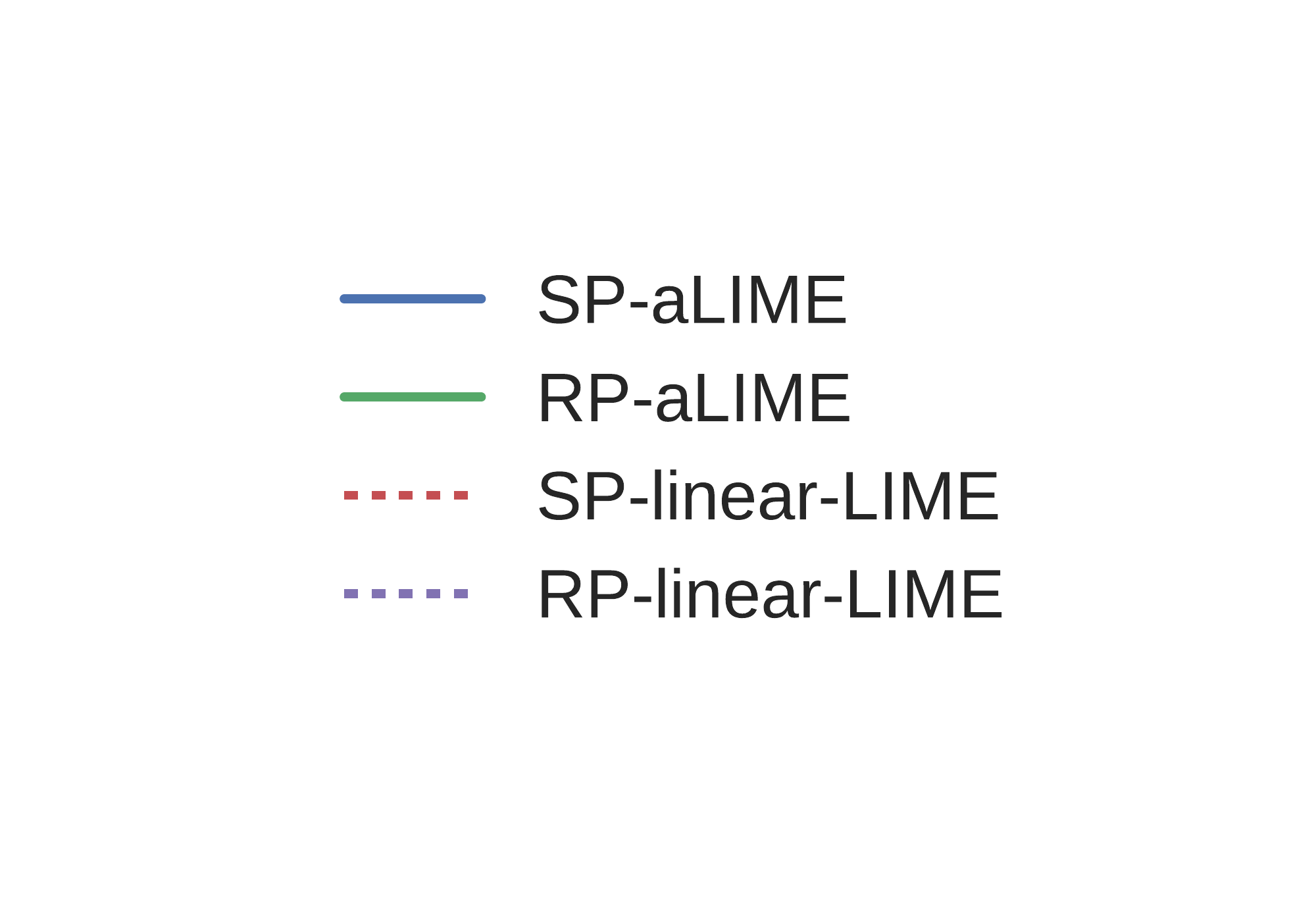}
  \end{subfigure}
  }
\label{diabetes-graphs}
\caption{Hospital Readmission dataset}
\end{figure}

We show precision-coverage plots of a single explanation ($K=1$) in Figures \ref{adultpreccov} and \ref{diabetespreccov}, where we vary $\tau$ for linear LIME and $\epsilon$ for aLIME. 
The results show that for any level of coverage, aLIME has better precision than linear LIME. Furthermore, using submodular pick greatly increases the coverage at the same precision level. Linear LIME performs particularly worse in the dataset with the highest number of dimensions (\emph{hospital readmission}), where the distance degrades.
We note that one of the main advantages of aLIME over linear LIME is making its coverage clear to humans - without human experiments, there is no way to know what these plots look like for linear LIME, but we can expect them to be the same for aLIME.

We vary the number of explanations the simulated user sees ($K$) in Figures \ref{adultprec}, \ref{adultcov}, \ref{diabetesprec} and \ref{diabetescov}.
In order to keep the results comparable, we set $\epsilon=0.05$ and picked $\tau$ such that the average precision at $K=1$ for linear LIME was at least $0.95$. 
In both datasets, aLIME is able to maintain higher precision regardless of how many explanations are shown, with coverage that dominates linear LIME.
It is worth noting that both datasets are of the same data type (tabular), and are such that the behavior of the model is simple in a large part of the input space (good conditions for aLIME), as demonstrated by the top $2$ anchors that maximize coverage for each dataset in Figure \ref{anchor-table}.
While these models are complex, their behavior on most of the input space (65\% for \emph{adult} and 81\% for \emph{hospital readmission}) is covered by these simple rules with high precision.

\presec
\begin{figure}
  \begin{subfigure}[t]{0.4\textwidth}
  \small
   \centering
  \begin{tabular}{|l|}
    \hline
    \textbf{IF} Education $\leq$ High School\\
    \textbf{THEN PREDICT} Salary $\leq 50K$ \\ 
    \\
    \textbf{IF} Marital status = Never married\\
    \textbf{THEN PREDICT} Salary $\leq 50K$ \\ \hline

  \end{tabular}
  \caption{Adult}
  \end{subfigure}
  \begin{subfigure}[t]{0.6\textwidth}
  \small
  \centering
  \begin{tabular}{|l|}
      \hline
    \textbf{IF} Inpatient visits = 0\\
    \textbf{THEN PREDICT} Never \\
    \\
    \textbf{IF} Inpatient visits $\geq 2$ \textbf{AND}
    Emergency visits $\geq 1$\\
    \hspace{3mm}\textbf{AND} Outpatient visits $\geq 1$\\
    \textbf{THEN PREDICT} $>30$ days \\
    \hline
  \end{tabular}
  \precap
  \caption{Hospital readmission}
  \end{subfigure}
  \caption{Top-2 anchors chosen with Submodular Pick for both datasets}
  \label{anchor-table}
\end{figure}

\begin{table}[tb]
  \small
  \begin{tabular}{lc|ll}
    \hline
    \multicolumn{2}{c|}{\bf Data and prediction} & \multicolumn{2}{c}{\bf Explanation} \\
    Sentence & Tag for word \emph{play} & \textbf{IF} & \textbf{THEN PREDICT} \\
    \hline
    I want to \textbf{play} ball. & VERB &  previous word is PARTICLE &  play is VERB. \\
    I went to a \textbf{play} yesterday. & NOUN &  previous word is DETERMINER & play is NOUN. \\
    I \textbf{play} ball on Mondays. & VERB & previous word is PRONOUN & play is VERB. \\
    \hline
  \end{tabular}
  \caption{Anchors for Part of Speech tagging}
  \postcap
  \label{spacy_table}
\end{table}

\begin{figure}[tb]
  \centering
  \begin{subfigure}[b]{0.3\textwidth}
    \centering
    \includegraphics[width=\textwidth]{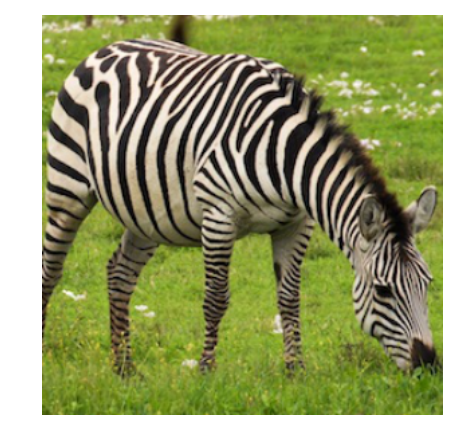}
  \caption{Original image}
  \end{subfigure}
  \begin{subfigure}[b]{0.3\textwidth}
    \centering
    \includegraphics[width=\textwidth]{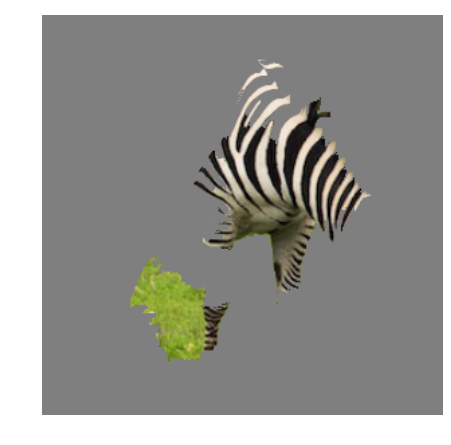}
  \caption{Anchor for ``Zebra''}
  \label{zebra-anchor}
  \end{subfigure}
  \begin{subfigure}[b]{0.32\textwidth}
    \begin{minipage}[b]{\textwidth}
        \includegraphics[width=0.45\textwidth]{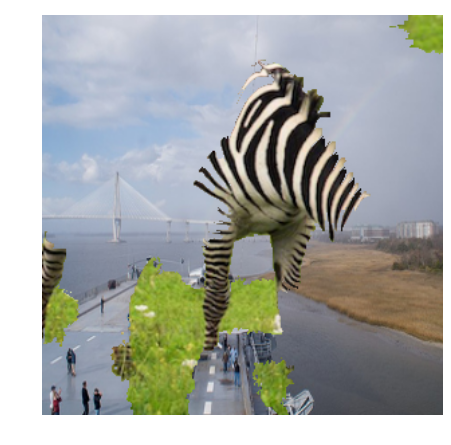}
        \includegraphics[width=0.45\textwidth]{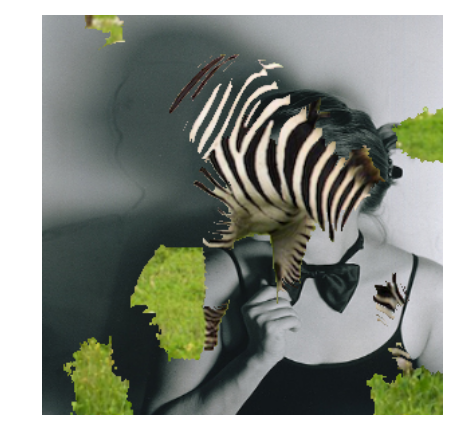}\\
        \includegraphics[width=0.45\textwidth]{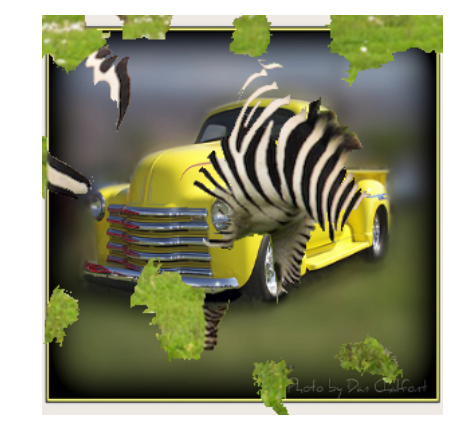}
        \includegraphics[width=0.45\textwidth]{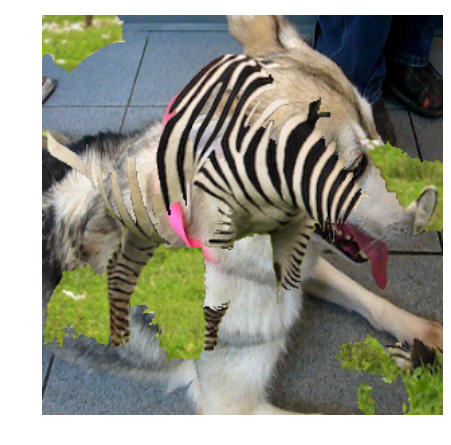}
    \end{minipage}
  \caption{Images with $P(\text{zebra}) > 90\%$}
  \label{zebra-others}
  \end{subfigure}
\caption{\textbf{Image classification}: explaining a prediction from Inception, and examples from $\Dist(z|c,x)$}
  \postcap
\label{zebra-all}
\end{figure}

\presec
\section{Qualitative Examples}
\postsec
\label{sec:case}

\textbf{Part-of-Speech tagging: }
We use a black box state-of-the-art POS tagger (\url{http://spacy.io}), and explain tag predictions for the word \emph{play} in different contexts in Table \ref{spacy_table}.
The anchors demonstrate that the POS picks up on the correct patterns. Furthermore, they are short and easy to understand.
Anchors are particularly suited for this task, where the dimensionality is small and the behavior of good models is more easily captured by IF-THEN rules than linear models.

\textbf{Image classification: }
We use aLIME to explain a prediction from the Inception V3 classifier on an image of a Zebra in Figure \ref{zebra-all}, where we first split the image into superpixels.
The anchor in Figure \ref{zebra-anchor} means that if we fix the non-grey superpixels, we can substitute the greyed-out superpixels by a random image, and the model will predict ``zebra'' around 95\% of the time. To illustrate this, we display on Figure \ref{zebra-others} a set of images from $\Dist(z|c,x)$ (i.e. where the anchor is fixed), and the model predicts ``zebra''.
While this choice of distribution produces images that look nothing like real images (Figure \ref{zebra-others}), it makes for more robust explanations than distributions that only hide parts of the image with gray or dark patches (\cite{goyal, lime}).
This anchor demonstrates that the model picks up on a pattern that does not require a zebra to have four legs, or even a head - which is a pattern very different than the patterns humans use to detect zebras.

\begin{figure}[h!]
    \centering
    \begin{subfigure}[b]{0.23\textwidth}
        \includegraphics[width=\textwidth]{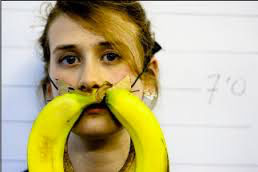}
        \caption{Original Image}
    \end{subfigure}
    \begin{subfigure}[b]{0.42\textwidth}
      \scriptsize
      \centering
      \begin{tabular}{lc}
        \toprule
          \color{purple}{\textbf{What} is the mustache made of?} & \color{purple}{banana} \\
        \midrule
        \textbf{What} is the ground made of ? & banana \\
        \textbf{What} is the bed made of ? & banana \\
        \textbf{What} is this mustache ? & banana \\
        \textbf{What} is the man made of? & banana \\
        \textbf{What} is the picture of ? & banana \\
        \bottomrule
      \end{tabular}
      \caption{}
      \label{qa1}
    \end{subfigure}
    \begin{subfigure}[b]{0.31\textwidth}
      \centering
      \scriptsize
      \begin{tabular}{lc}
        \toprule
        \color{purple}{How \textbf{many} bananas are in the picture?}  & \color{purple}{2} \\
        \midrule
        How \textbf{many} are in the picture?& 2 \\
        \textbf{many} animals the picture ?    & 2 \\
        How \textbf{many} people are in the picture ? & 2 \\
        How \textbf{many} zebras are in the picture ? & 2 \\
        How \textbf{many} planes are on the picture ? & 2 \\
        \bottomrule
      \end{tabular}
      \caption{}
      \label{qa2}
    \end{subfigure}
\caption{\textbf{Visual QA:} explaining predictions from a CNN-LSTM model by looking at the question text (image is fixed), and examples from $\Dist(z|c, x)$ }
\label{vqa-figure}
\end{figure}

\textbf{Visual Question Answering: }
Visual QA \cite{vqa} models are multi-modal, and thus can be explained in terms of the image, the question or both.
Here, we find anchors on the questions, leaving the image fixed, and use a bigram language model trained on input questions as $\Dist$.
%
We select two questions to explain, which are the top rows (in purple) of Figures \ref{qa1} and \ref{qa2}. The anchors (in bold), are respectively ``What'' and ``many'', and we show questions drawn from $\Dist(z | c, x)$ below the original question.
%
The first anchor states that if ``What'' is in the question, the answer will be ``banana'' about 95\% of the time, while the latter states the same about ``many'' and ``2'', respectively
-- both explanations clearly indicate undesirable behavior from the model.
Again, this kind of explanation is intuitive and easier to understand than a linear model, even one with high weight on the words ``What'' and ``banana'', as one knows exactly when it applies and when it does not.

\newpage
\section{Conclusion}
In this work, we argued that high precision and clear coverage bounds are very desirable properties of model-agnostic explanations.
We introduced aLIME, a system is designed to produce rule-based explanations that exhibit both these properties.
IF-THEN rules are intuitive and easy to understand, and identifying parts of the input that result in prediction invariance (i.e. the rest does not matter) is similar to how humans explain many of their choices.
We demonstrated aLIME's flexibility by explaining predictions from a variety of classifiers on a myriad of domains, outperforming linear explanations from LIME on simulated experiments.

\bibliographystyle{plainnat}
\bibliography{references}

\end{document}